\pdfoutput=1

\documentclass[11pt]{article}

\usepackage[final]{acl}

\usepackage{times}
\usepackage{latexsym}

\usepackage[T1]{fontenc}

\usepackage[utf8]{inputenc}

\usepackage{microtype}

\usepackage{inconsolata}

\usepackage{graphicx}
\usepackage{subcaption}

\usepackage{booktabs}
\usepackage{tabularx} 
\usepackage{caption}   

\title{Does Acceleration Cause Hidden Instability in Vision Language Models? \\ Uncovering Instance-Level Divergence Through a \\ Large-Scale Empirical Study}

\author{
 \textbf{Yizheng Sun\textsuperscript{1}}\quad
 \textbf{Hao Li\textsuperscript{1}}\quad
 \textbf{Chang Xu\textsuperscript{2}}\quad
 \textbf{Hongpeng Zhou\textsuperscript{1}} \\
 \textbf{Chenghua Lin\textsuperscript{1}}\quad
 \textbf{Riza Batista-Navarro\textsuperscript{1}}\quad
 \textbf{Jingyuan Sun\textsuperscript{1,*}}
\\
 \textsuperscript{1}University of Manchester\quad
 \textsuperscript{2}Microsoft Research
\\
 \small{
   \textbf{*Correspondence:} \href{jingyuan.sun@manchester.ac.uk}{jingyuan.sun@manchester.ac.uk}
 }
}

\begin{document}
\maketitle
\begin{abstract}
Vision-Language Models (VLMs) are powerful yet computationally intensive for widespread practical deployments. To address such challenge without costly re-training, post-training acceleration techniques like quantization and token reduction are extensively explored. However, current acceleration evaluations primarily target minimal overall performance degradation, overlooking a crucial question: does the accelerated model still give the same answers to the same questions as it did before acceleration? This is vital for stability-centered industrial applications where consistently correct answers for specific, known situations are paramount, such as in AI-based disease diagnosis. 
We systematically investigate this for accelerated VLMs, testing four leading models (LLaVA-1.5, LLaVA-Next, Qwen2-VL, Qwen2.5-VL) with eight acceleration methods on ten multi-modal benchmarks. Our findings are stark: despite minimal aggregate performance drops, accelerated models changed original answers up to 20\% of the time. Critically, up to 6.5\% of these changes converted correct answers to incorrect. Input perturbations magnified these inconsistencies, and the trend is confirmed by case studies with the medical VLM LLaVA-Med. This research reveals a significant oversight in VLM acceleration, stressing an urgent need for instance-level stability checks to ensure trustworthy real-world deployment.
\end{abstract}

\section{Introduction}
\label{sec:introduction}
Large Vision-Language Models (VLMs) are demonstrating remarkable capabilities in understanding and generating content across visual and textual modalities \citep{llava1.5,liu2024llavanext,qwen2.5-vl,sun2024lanvikd}.
\begin{figure}[t]
\centering
\includegraphics[width=0.48\textwidth]{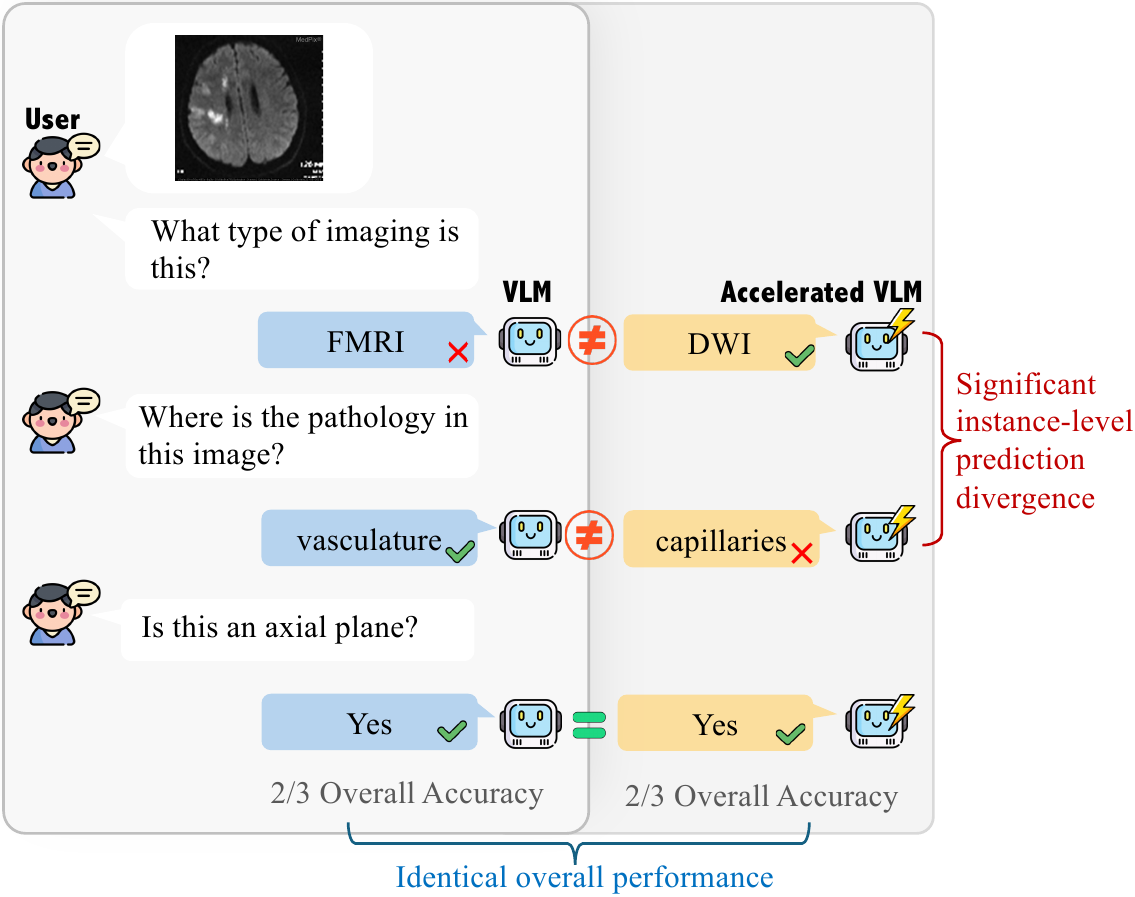}
\caption{\label{intro_pic} Current VLM acceleration methods focus on improving efficiency while minimizing overall performance drop relative to the base model. However, this focus may obscure a critical risk: accelerated models can exhibit significant changes in instance-level predictions compared to their original counterparts. Such instability poses serious concerns in sensitive domains such as healthcare, where producing stable and reliable outputs is essential.}
\end{figure}
Despite their impressive performance, the substantial computational demands of state-of-the-art VLMs critically limit their practical deployment, particularly in resource-constrained environments \citep{FastV, llava-mini, bandwidth, li2025bridge}. To mitigate these challenges without the necessity of costly re-training, post-training acceleration techniques—such as quantization \citep{awq, gptq, llmint8} and token reduction \citep{FastV, visionzip,pyramidDrop,sun2025silent,sun2025lvpruning}—are widely adopted. The primary objectives of these techniques have been two-fold: achieving substantial computational efficiency gains while ensuring minimal degradation in aggregate performance metrics. Yet, this prevailing focus obscures other vital impacts of acceleration, posing the question: Are these two criteria truly sufficient to guarantee the reliable deployment of accelerated VLMs in practice?

We contend that for many practical applications, particularly in critical domains like medicine \citep{BioGPT, llava-med}, the answer is highly risky to be ``No''. In such fields, system development and validation often adhere to a ``case-driven'' paradigm \citep{case-driven0,case-driven1,case-driven2}, where a fundamental requirement is the AI system's ability to consistently and correctly resolve specific, known crucial instances, even post-optimization or updates. Consider a medical VLM adept at identifying a rare disease from patient scans; it is paramount that this specific diagnostic capability remains invariant after an acceleration process aimed at enhancing efficiency. However, as illustrated in Figure \ref{intro_pic}, this crucial aspect of instance-level stability is largely unaddressed within the evaluation of current acceleration methodologies \citep{awq, gptq, FastV, visionzip}, potentially masking significant operational risks.

This paper confronts this oversight by systematically investigating the instance-level stability of accelerated VLMs. Our central aim is to evaluate whether and to what extent existing post-training acceleration techniques, despite ostensibly preserving overall performance, can induce substantial and often detrimental inconsistencies in models' response to individual inputs. To rigorously quantify this instability, we introduce two intuitive yet powerful metrics: Divergence Ratio (DR) and Negative Divergence Ratio (NDR). DR measures the frequency with which an accelerated model yields a different prediction for the same input compared to its original, unaccelerated counterpart. NDR quantifies a more critical failure mode: the proportion of instances where a correct prediction from the original model becomes incorrect after acceleration. Low DR and NDR values signify that an accelerated VLM maintains behavioral fidelity and reliability. Conversely, high values—even when accompanied by negligible shifts in aggregate performance—would indicate that the accelerated model's behavior has become alarmingly unpredictable relative to its original state.

To empirically validate our hypothesis, we undertook an extensive study. We assessed eight distinct acceleration methods applied to four leading open-source VLMs (LLaVA-1.5 \citep{llava1.5}, LLaVA-Next \citep{liu2024llavanext}, Qwen2-VL \citep{Qwen2-vl}, and Qwen2.5-VL \citep{qwen2.5-vl}) across ten diverse multi-modal benchmarks. To probe the resilience of instance-level stability under practical conditions, we further evaluated model performance on perturbed inputs (spanning both visual and textual modalities) designed to mimic real-world data variations. Underscoring the high stakes involved, we conducted targeted case studies on LLaVA-Med \citep{llava-med}, a VLM tailored for medical applications where predictive consistency is non-negotiable. Our experiments reveal several striking findings:

\begin{enumerate}
    \item Despite acceleration methods inducing only a negligible drop in overall performance (average of 0.8\%), they precipitated surprisingly high Divergence Ratios (DR) of up to 20\% and, more critically, Negative Divergence Ratios (NDR) reaching up to 7\%.
    \item Input data perturbations, characteristic of real-world scenarios, further exacerbated this divergence.
    \item Application of acceleration to the medical VLM (LLaVA-Med) corroborated these high DR and NDR values, highlighting the acute potential risks in safety-critical domains.
\end{enumerate}
To the best of our knowledge, this work represents the first large-scale empirical investigation dedicated to the instance-level stability of VLM acceleration techniques. Our research uncovers a significant, potentially hazardous, oversight in current VLM acceleration practices, emphasizing an urgent imperative for incorporating rigorous instance-level stability checks to ensure these models are genuinely faithful and trustworthy for real-world deployment.

\section{Related Work}
\subsection{Large Vision-Language Models}
Large Vision-Language Models (VLMs) have advanced rapidly in integrating visual and textual understanding. Early models like CLIP \citep{clip} employed contrastive learning to align these modalities. Subsequent architectures, such as BLIP-2 and Instruct-BLIP \citep{blip2, instructBLIP}, introduced Q-Former to bridge pre-trained vision encoders with Large Language Model (LLM) backbones. More recent state-of-the-art models, including LLaVA-1.5 \citep{llava1.5}, LLaVA-NeXT \citep{liu2024llavanext}, and the Qwen-VL series \citep{Qwen2-vl, qwen2.5-vl}, leverage powerful LLMs (e.g., Vicuna, LLaMA, Qwen2 \citep{vicuna, llama3, qwen2}) and lightweight vision-text connectors (typically linear layers) for advanced multimodal reasoning. However, VLM vision encoders often generate a high volume of visual tokens (hundreds or thousands \citep{clip}). The LLM backbone processing of these numerous tokens incurs significant computational costs, hindering the practical deployment of VLMs.

\subsection{Post-Training Acceleration Techniques for Vision-Language Models}
Post-training acceleration techniques are widely applied to reduce computational demands of VLMs without costly retraining. Token Reduction methods aims to substantially remove the redundant visual tokens for VLMs, thereby reducing the input sequence length and lowering inference costs. Recent methods implementing this approach during inference include VisionZip \citep{visionzip}, PyramidDrop \citep{pyramidDrop}, FastV \citep{FastV}, SparseVLM \citep{SparseVLM}, and HiRed \citep{Hired}. Quantization techniques reduces model size and computational overhead by utilizing lower-precision numerical formats (e.g., 8-bit, 4-bit) for model weights and/or activations. Post-Training Quantization (PTQ), which applies this technique after model training, has become a common practice, such as LLM.int8() \citep{llmint8}, GPTQ \citep{gptq}, and AWQ \citep{awq}. Although these methods often report minimal degradation on standard benchmarks, their impact on instance-level stability remains largely unexplored. This work systematically investigates the instance-level prediction stability of VLMs under both token reduction and quantization, moving beyond standard benchmark evaluations.

\subsection{Evaluation for LM Acceleration}
The typical approach to evaluating model acceleration techniques tends to emphasize negligible loss in aggregate performance and improved computational efficiency. However, there's a growing recognition that such criteria, while important, may overlook other critical impacts. Recent investigations, for example, have shown that quantization can diminish the reasoning capabilities of LLMs \citep{quant_meet_reasoning}, and that prompt compression can affect their ability to retain information \citep{prompt_compress}. Similarly, \citet{acc_not} demonstrates that accuracy alone is not enough for assessing LLM quantization, leading to proposals like the ``flip'' metric for instance-level changes. \citet{DBLP:journals/corr/abs-2502-11501} argues that the fundamental designs of token reduction methods for VLMs can cause biased performance on different task types. Moreover, a specialized benchmark, LLMCBench, has been introduced targeting the practical efficiency of model compression techniques for real-world deployment \citep{DBLP:conf/nips/YangHGWDLQJ024}. Distinct from these explorations, our work concentrates on a crucial aspect: the instance-level stability and reliability of accelerated VLMs, ensuring they consistently solve the problems they were initially capable of solving.

\section{Experimental Settings}

\subsection{Tasks and Datasets}
We utilize a diverse suite of ten benchmark datasets covering various Visual-Language understanding capabilities. These include AI2D \citep{ai2d} for diagram understanding, GQA \citep{gqa} for real-world compositional reasoning, MMBench \citep{mmbench} for diverse multi-modal abilities, MMMU \citep{yue2023mmmu} for expert-level multi-discipline reasoning, OK-VQA \citep{ok_vqa} requiring external knowledge, POPE \citep{pope} for evaluating object hallucination, ScienceVQA \citep{scienceqa} focusing on science diagrams, TextVQA \citep{textvqa} requiring reading text within images, VizWiz \citep{vizwiz} using images from visually impaired users, and the widely-used large-scale VQA benchmark VQAv2 \citep{balanced_vqa_v2}. Finally, we use VQA-RAD \citep{vqarad} to extend to medical domain tasks.  Details of the benchmarks are presented in Appendix \ref{appen: benchmarks}.

\subsection{Base Models and Acceleration Techniques}
We select four state-of-the-art open-source VLMs as base models for our acceleration experiments. LLaVA-1.5 \citep{llava1.5} is a widely recognized VLM demonstrating strong general vision-language capabilities. LLaVA-Next \citep{liu2024llavanext} extends LLaVA-1.5, improving performance particularly for high-resolution inputs. Qwen2-VL \citep{Qwen2-vl} and Qwen2.5-VL \citep{qwen2.5-vl} are recent released VLMs, which are adept at handling various image resolutions and video inputs. Additionally, we also use LLaVA-Med \citep{llava-med}, which is a specialised medical domain VLM. We adopt the 7B model size for all VLMs throughout our study, unless stated otherwise. 

We investigate two main categories of post-training acceleration: token reduction and quantization. For token reduction, we evaluate five of the latest and widely applied methods, including VisionZip \citep{visionzip}, which selects informative tokens and merges others; PyramidDrop \citep{pyramidDrop}, which progressively drops tokens in deeper layers; SparseVLMs \citep{SparseVLM}, which prunes tokens based on relevance scores; FastV \citep{FastV}, dynamically pruning based on attention scores during inference; and HiRed \citep{Hired}, designed for high-resolution inputs, allocating token budgets based on attention. For all the token reduction methods, we choose the signature or best-performing hyper-parameter settings as reported in the corresponding papers, which are listed in Appendix \ref{appen:hyper-param}. For Quantization, which reduces numerical precision, we apply: llm.int8() \citep{llmint8} (W8A16), a mixed-precision quantisation scheme; AWQ \citep{awq} (W4A16), an activation-aware 4-bit weight quantization; and GPTQ \citep{gptq} (W4A16), a layer-wise 4-bit weight quantization method.

\subsection{Evaluation Metrics}
We report standard top-1 accuracy for all tasks except for POPE, where F1 score is the standard metric. We also calculate the Accuracy or F1 Drop for all the acceleration methods compared with the corresponding baseline VLMs. To assess the instance-level instability of accelerated models compared to their original counterparts, we introduce two additional metrics: \textbf{1) Divergence Ratio (DR)}, defined as the proportion of test samples where the accelerated model's prediction differs from the original model's prediction, irrespective of correctness. \textbf{2) Negative Divergence Ratio (NDR)}, which quantifies harmful instability by measuring the proportion of samples that were correctly predicted by the original model but incorrectly predicted by the accelerated model.

\subsection{Input Perturbation}
To better understand the instance-level stability of accelerated VLMs under practical settings, we adopt a comprehensive set of input perturbation methods proposed by \citet{perturbation} to simulate the real-world user scenarios. Specifically, we use 96 visual perturbation methods (e.g. noise, blur, weather effects) and 87 textual perturbation methods (e.g., typos, paraphrasing, character substitutions), whose details are shown in Appendix \ref{appen:perturbation_methods}. We apply these visual and textual perturbations separately to the inputs of the accelerated models and assess their impact on performance and prediction stability.

\begin{table*}[t]
\centering
\includegraphics[width=0.97\textwidth]{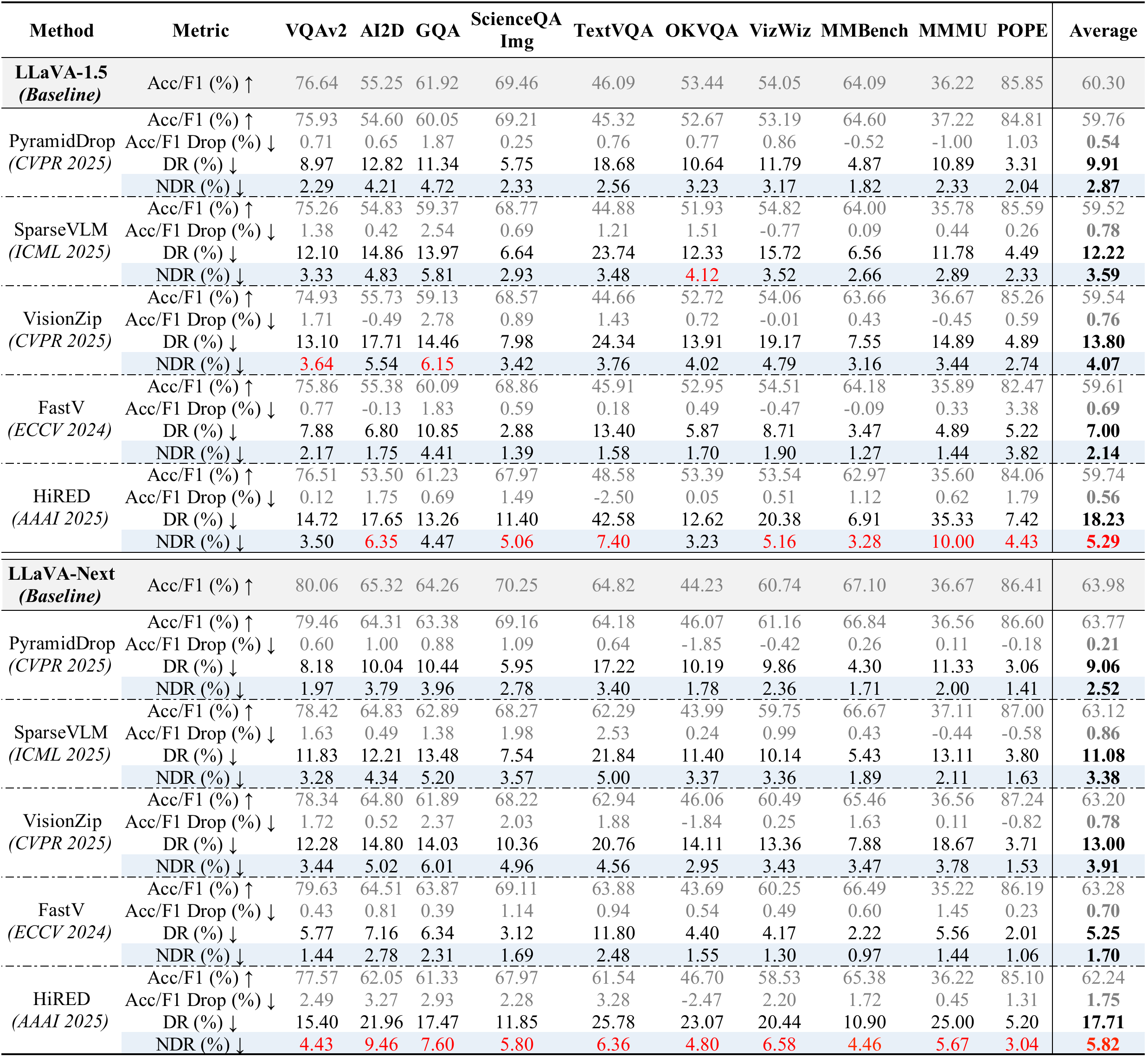}
\caption{\label{tab:token_reduction_res} \textbf{Instance-Level Instability in Token Reduction Methods.} For each acceleration method, we report: Accuracy (Acc) for most benchmarks (F1 score for POPE \citep{pope}), Acc/F1 drop (performance degradation vs. baseline), Divergence Ratio (DR), and Negative Divergence Ratio (NDR) to evaluate instance-level prediction changes. Red values indicate the largest NDR per baseline model within each benchmark column. Across all benchmarks and token reduction methods, results reveal high DR and NDR values despite negligible Acc/F1 drops, signifying considerable instance-level prediction instability.}
\end{table*}

\begin{table*}[t]
\centering
\includegraphics[width=0.97\textwidth]{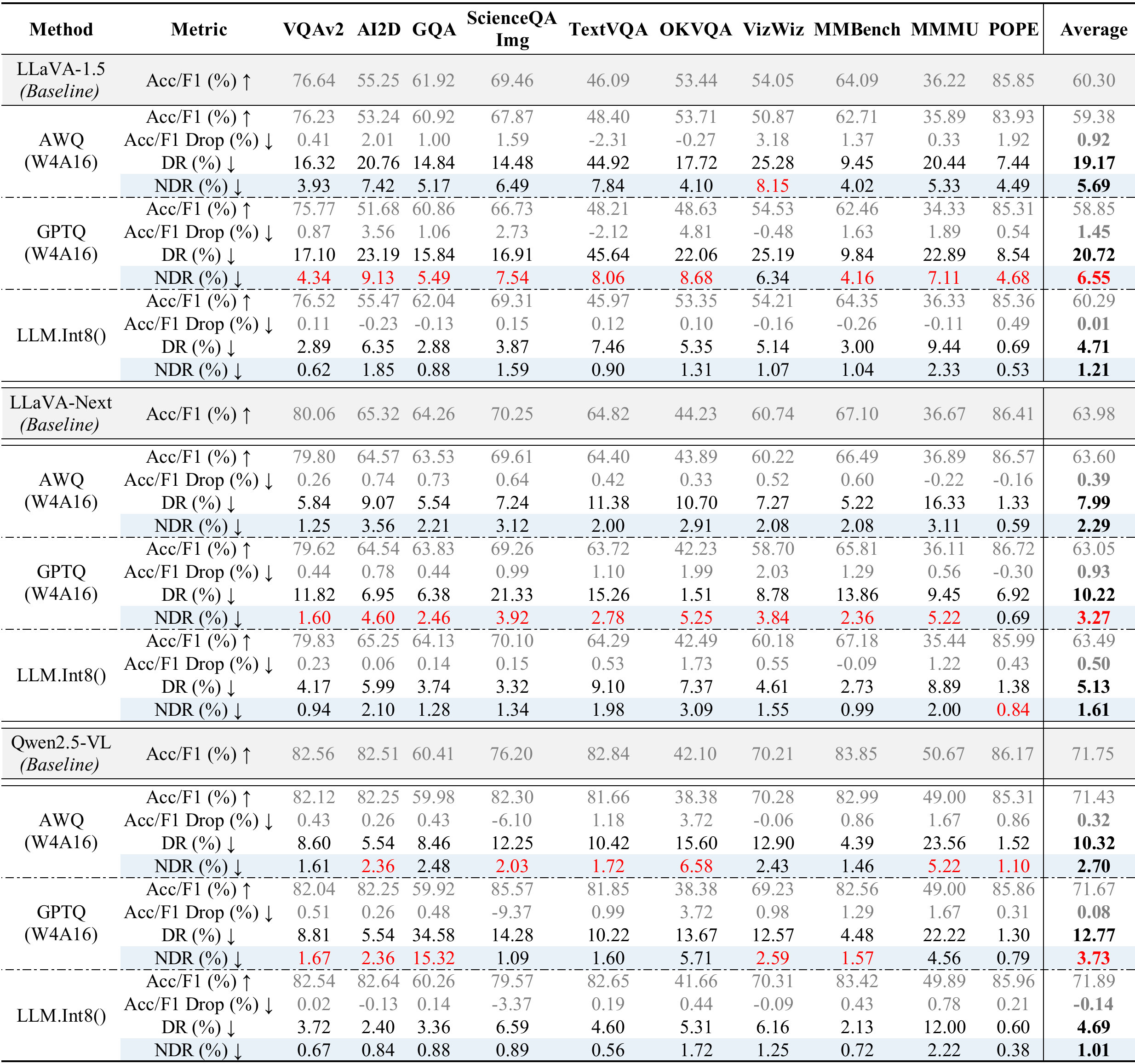}
\caption{ \label{tab:quant_res} \textbf{Instance-Level Instability in Quantization Methods.} This table presents Acc/F1, Acc/F1 Drop, DR, and NDR for various quantization methods. Most methods exhibit high DR and NDR values, indicating significant instance-level instability, similar to token reduction techniques. Only the LLM.int8() method \citep{llmint8} is a notable exception, maintaining relatively low DR and NDR. Red values indicate the largest NDR per baseline model within each benchmark column.}
\end{table*}

\section{Experimental Results}

This section presents our empirical findings on the instance-level stability of accelerated Vision-Language Models (VLMs). Our experiments are structured in three stages:

\begin{enumerate}
    \item We first evaluate Divergence Ratios (DR) and Negative Divergence Ratios (NDR) for widely used post-training acceleration methods (Token Reduction and Quantization) on standard benchmarks in section \ref{sec:standard_empirical_results}. This establishes their fundamental impact on instance-level stability under laboratory conditions.

    \item Next, we further assess the instance-level stability under more realistic conditions by applying input perturbations to large-scale Visual Question Answering (VQA) benchmarks (VQAv2 \citep{balanced_vqa_v2} and GQA \citep{gqa}), simulating typical input noise encountered in practice as discussed in section \ref{sec:perturbation_empirical_results}.

    \item Finally, we analyze an accelerated medical VLM to demonstrate the potential downstream consequences and critical risks of instance-level instability in a high-stakes domain in section \ref{sec:medical_results}.
\end{enumerate}


\subsection{Instance-Level Instability on Standard Benchmarks}
\label{sec:standard_empirical_results}

This section presents our quantitative findings on the instance-level stability of various post-training acceleration techniques applied to leading Vision-Language Models (VLMs). The detailed results for token reduction techniques are shown in Table~\ref{tab:token_reduction_res} and those for quantization methods are summarized in Table~\ref{tab:quant_res}. Qualifying examples are demonstrated in appendix \ref{appen:qualifying_examples}.

\paragraph{The Illusion of Stability: High Divergence Despite Low Aggregate Performance Drops.}
The most striking revelation from our experiments is the significant instance-level instability introduced by many common acceleration methods, even when these methods exhibit only minimal degradation in overall aggregate performance. This creates an illusion of stability if one only considers coarse-grained metrics. Across multiple VLMs and benchmarks, we consistently observed that \textbf{accelerated models altered their original predictions on identical inputs up to 20\% of the time (DR)}, a concerning level of divergence. More critically, our findings indicate that \textbf{up to 6.5\% of these changes converted previously correct answers into incorrect ones (NDR)}, directly undermining the model's reliability on specific, previously solved cases.

\paragraph{Instance-Level Instability in Token Reduction Methods.}
\label{subsec:token_reduction_instability}

Our investigation into token reduction techniques reveals substantial instance-level instability (Table~ \ref{tab:token_reduction_res}). The HIRED method, for example, when applied to LLaVA-1.5 and LLaVA-Next, caused minimal average aggregate performance drops ($\sim$0.2-0.6\%) but still led to high average DRs of $\sim$18\% and average NDRs approaching 6\%. Specific benchmarks under this method saw NDRs reach up to 9-10\% and DRs over 25\%. Other token reduction techniques like VisionZip and SparseVLM similarly produced notable DR and NDR values (e.g., average DRs often exceeding 12-13\%) despite their modest impact on overall accuracy scores. Since the Qwen-VL model series \citep{Qwen2-vl,qwen2.5-vl} already features integrated token compression modules, we do not separately evaluate the impact of external token reduction methods.

\paragraph{Instance-level Instability in Quantization Methods.}
\label{subsec:quant_instability}

The phenomenon of high instance-level instability extends to quantisation methods as shown in table \ref{tab:quant_res}. For instance, aggressive W4A16 quantization methods like GPTQ and AWQ applied to LLaVA-1.5 resulted in average aggregate performance drops of only $\sim$0.9-1.5\%, yet induced high average Deviation Ratios (DR) of $\sim$19-21\% and average Negative Deviation Ratios (NDR) of $\sim$5.7-6.6\%. Individual benchmarks exhibited even more severe divergence, with DRs occasionally exceeding 40\% and NDRs surpassing 8\%. While less aggressive techniques like LLM.int8() showed markedly lower DR/NDR values (e.g., LLaVA-1.5 average DR 4.71\%, NDR 1.21\%), the trend for commonly used aggressive quantization is a significant and concerning level of instance-level prediction change. Table \ref{tab:quant_res} only includes the results of Qwen2.5-VL \citep{qwen2.5-vl} for the Qwen-VL model series since it is the improved version of Qwen2-VL. We show the results of Qwen2-VL \citep{Qwen2-vl} separately in Appendix \ref{appen:qwen2-vl}.

In summary, these results underscore a critical, largely overlooked deficiency in current VLM acceleration practices. To better view the overall distribution of relation between Acc/F1 Drop and DR/NDR values, we visualize the data in Appendix \ref{appen:data_visual}. \textbf{The substantial DR and NDR values with minimal changes in aggregate metrics, provide compelling evidence that accelerated models can indeed become unreliable for specific instances they previously handled correctly.}

\subsection{Instance-Level Instability Under Input Perturbations}
\label{sec:perturbation_empirical_results}
\begin{table}[t]
    \centering
    \includegraphics[width=0.48\textwidth]{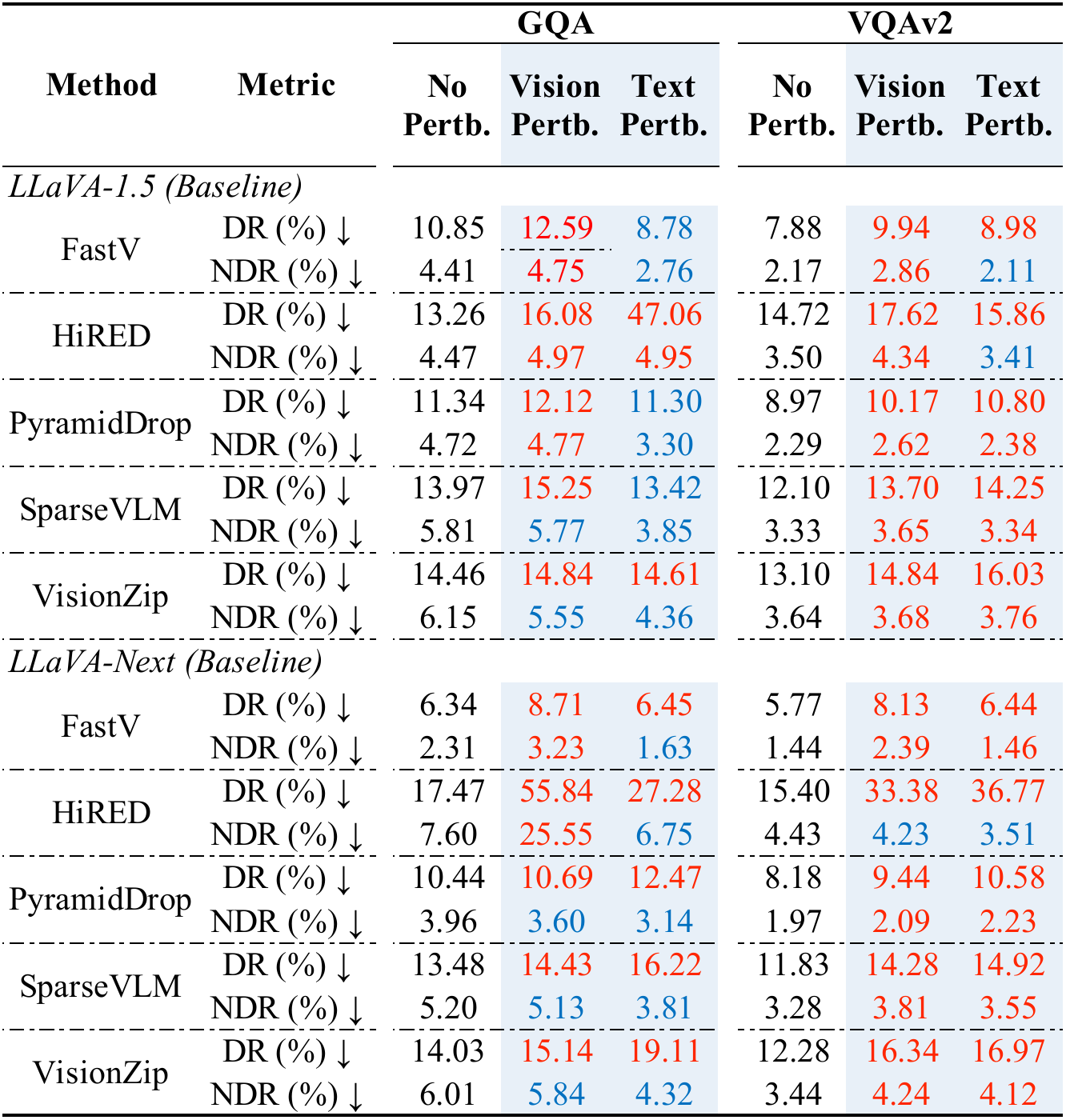}
    \caption{\label{token_reduction_perturb} Instance-level instability of token reduction methods under input perturbation. This table reports Divergence Ratio (DR) and Negative Divergence Ratio (NDR) across three input states: ``No Pertb.'' (original inputs), ``Vision Pertb.'' (e.g., image noise, blur), and ``Text Pertb.'' (e.g., text misspellings, paraphrasing). Red signifies higher DR/NDR under perturbation than without; blue signifies lower. The table illustrates that most methods suffer greater instance-level instability when inputs are perturbed.}
\end{table}

\begin{table}[ht]
    \centering
    \includegraphics[width=0.48\textwidth]{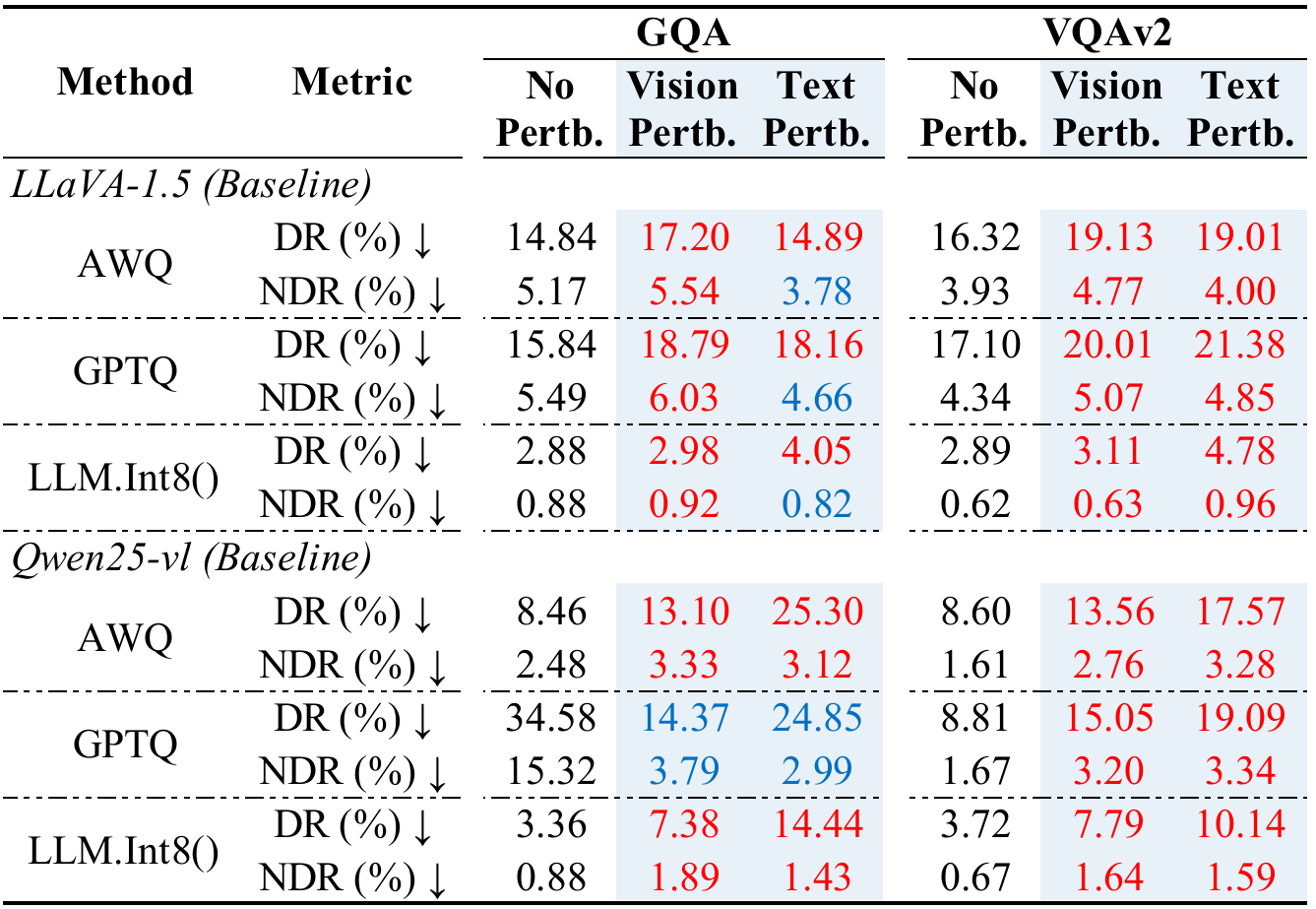}
    \caption{\label{quant_perturb}Instance-level instability of quantisation methods under input perturbation. Most quantisation methods demonstrate increased instance-level instability under input perturbations.}
\end{table}

To further demonstrate the risk of instance-level instability under practical settings, we conducted experiments involving perturbations to both text and vision inputs to VLMs, representing common real-world inputs disturbances.
The detailed results are shown in Table \ref{token_reduction_perturb} and Table \ref{quant_perturb}. We only show the DR and NDR values in the tables. Acc/F1 and Acc/F1 Drop values are listed in appendix \ref{appen:perturbation_methods}.  \textbf{The clear takeaway is that these perturbations generally exacerbate the Divergence Ratios (DR) and Negative Divergence Ratios (NDR) already observed in non-perturbed conditions.} For instance, applying vision perturbation to LLaVA-1.5 with AWQ quantization on VQAv2 increased its DR from 16.32\% to 19.13\% and its NDR from 3.93\% to 4.77\%. Text perturbation on the same model and benchmark also increased DR to 19.01\% and NDR, albeit slightly, to 4.00\%. Similarly, for token reduction, LLaVA-1.5 with the HIRED method on GQA saw vision perturbation elevate DR from 13.26\% to 16.08\% and NDR from 4.47\% to 4.97\%; text perturbation in this case markedly increased DR to 47.06\% and NDR to 4.95\%. This observed pattern of increased instability under noisy conditions was generally consistent across different types of acceleration methods, including both quantization and token reduction. \textbf{Consequently, the levels of instance-level instability likely aggravate when these accelerated models are deployed in dynamic, real-world environments where input data is rarely pristine.}

\subsection{Instance-Level Prediction Instability in the Medical Domain}

In this section,  we apply VisionZip \citep{visionzip}, PyramidDrop \citep{pyramidDrop}, and LLM.int8() \citep{llmint8} to LLaVA-Med \citep{llava-med}. We firstly verify the generalisation of these acceleration methods by evaluating them on the biomedical multimodal conversation test set introduced by \citet{llava-med}. We then conduct a case study by measuring the DR and NDR values on a medical VQA dataset VQA-RAD \citep{vqarad}, revealing similarly high DR and NDR values as shown on general domain benchmarks as discussed in Section \ref{sec:standard_empirical_results}.

\begin{table*}[t]
\centering
\includegraphics[width=0.95\textwidth]{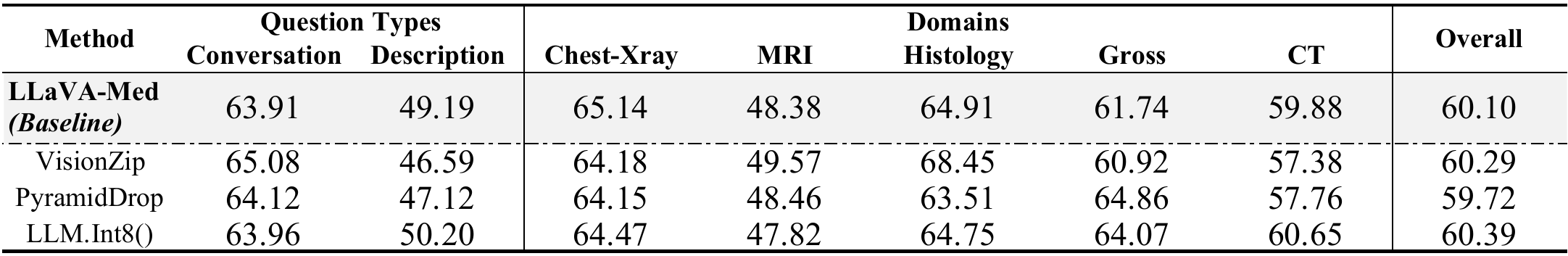}
\caption{Evaluation of VisionZip \citep{visionzip}, PyramidDrop \citep{pyramidDrop}, and LLM.int8() \citep{llmint8} applied to LLaVA-Med \citep{llava-med} on its biomedical multimodal conversation test set. The results confirm the negligible overall performance impact of extending these acceleration techniques to the medical domain.}
\label{tab:medical_performance}
\end{table*}
\begin{table}[t]
    \centering
    \includegraphics[width=0.4\textwidth]{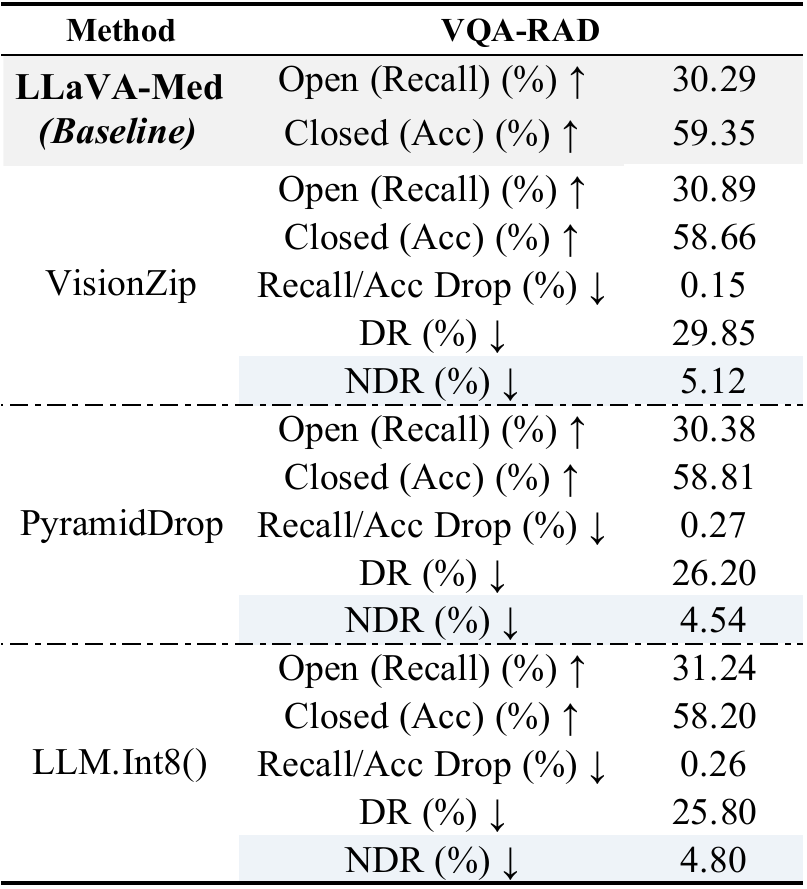}
    \caption{\label{vqarad} 
    Evaluation of VisionZip \citep{visionzip}, PyramidDrop \citep{pyramidDrop}, and LLM.int8() \citep{llmint8} on LLaVA-Med \citep{llava-med} using the VQA-RAD \citep{vqarad} dataset (comprising open-ended and closed-ended questions). While aggregate performance loss was minimal, all three acceleration methods exhibited significant instance-level deviations.}
\end{table}

\paragraph{Generalisation of Acceleration Methods to Medical Domain.}
\label{sec:medical_results}
Table ~\ref{tab:medical_performance} summarizes the performance of various acceleration methods compared to the baseline model LLaVA-Med on the biomedical multimodal conversation test set. Results indicate that all examined acceleration methods (VisionZip, PyramidDrop and LLM.int8()) maintained almost identical performance to the baseline across diverse medical imaging modalities. This demonstrates minimal overall performance impact from generalising acceleration methods to medical context.

\paragraph{High Risk Instance-Level Instability in Medical Domain.}
Despite minimal overall performance drop, significant instance-level deviations were observed on the VQA-RAD benchmark as shown in Table \ref{vqarad}. Deviation Ratio (DR) values were notably high, ranging between 25.80\%-29.85\% across the evaluated methods, suggesting that accelerated models frequently altered their predictions compared to the baseline model. More critically, Negative Deviation Ratios (NDR), representing detrimental prediction changes, were considerably higher in the medical domain (4.54\%-5.12\%) compared to general domain benchmarks. This indicates heightened instability risks when deploying accelerated VLMs in high-stake medical applications, where unstable outputs such as misdiagnoses could have severe consequences.

\section{Conclusion}

We presented a large-scale empirical study of instance-level stability in accelerated Vision–Language Models. Across four VLMs, eight post-training acceleration methods, and ten benchmarks, we found substantial \emph{Divergence Ratios} (DR; up to 20\%) and non-trivial \emph{Negative Divergence Ratios} (NDR; up to 6.5\%) despite negligible changes in aggregate accuracy/F1. Instability increased under realistic input perturbations and was corroborated in a medical VLM, highlighting concrete risks for safety-critical applications.

These findings show that aggregate metrics alone are insufficient for assessing the reliability of accelerated VLMs. We therefore recommend: (i) reporting DR and NDR alongside standard metrics; (ii) evaluating under documented perturbation regimes; and (iii) incorporating targeted, case-driven tests for critical instances prior to deployment. Going forward, we will extend evaluations to real-world industrial datasets and workloads, longitudinal settings with natural drift, and end-to-end deployment studies to establish external validity and to guide stability-aware acceleration strategies.

\section{Limitations}

This study is subject to several limitations that qualify the interpretation and generalizability of the results. First, the empirical analyses rely predominantly on synthetic data and curated academic benchmarks evaluated under controlled laboratory conditions. Although input-perturbation protocols were employed to approximate real-world variability, such simulations are, at best, partial surrogates for the heterogeneity, nonstationarity, and operational constraints observed in practice. As a result, the evidence concerning instance-level instability should be regarded as indicative rather than definitive for production contexts. Second, the proposed approach has not yet been evaluated on real-world or industrial test cases; accordingly, claims of external and ecological validity remain provisional. We therefore caution against direct extrapolation of the reported quantitative estimates to domains with domain-specific requirements. Future work will prioritize rigorous assessments on representative industrial datasets and workloads, longitudinal evaluations under naturally occurring drift, and end-to-end deployment studies to substantiate and refine these findings.

\bibliography{main}

\appendix

\section{Benchmark Details}
\begin{table}[ht] 
\centering
\includegraphics[width=0.44\textwidth]{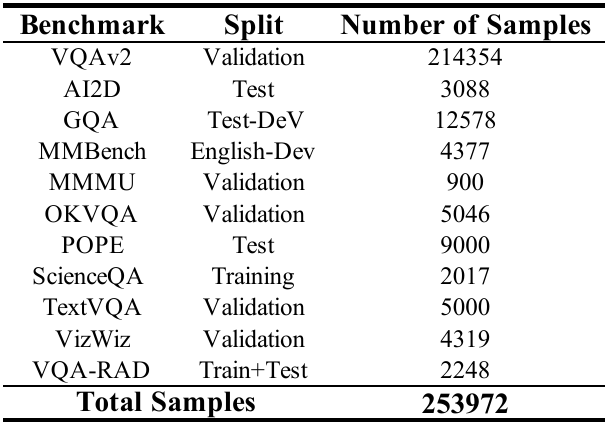}
\caption{Summary of benchmark datasets, splits, and their respective sample sizes.}
\label{tab:benchmarks_detailed}
\end{table}
\label{appen: benchmarks}
To comprehensively evaluate our methods, we utilize a diverse array of ten established benchmarks, as detailed in Table \ref{tab:benchmarks_detailed}. This selection spans various visual and multimodal understanding tasks, including Visual Question Answering (VQAv2, GQA, AI2D, OKVQA, TextVQA, ScienceQA, VizWiz), multimodal reasoning (MMMU), and general multimodal capabilities (MMBench, POPE). The evaluation is conducted on standard splits such as validation, test, or development sets, encompassing a significant total of 251,679 samples. Notably, VQAv2 contributes the largest portion with 214,354 validation samples, ensuring a robust assessment across different challenge domains and scales. For evaluation in the medical domain, we utilize the VQA-RAD benchmark, employing both its training and test sets. This dataset comprises 1299 closed-ended (yes/no) questions, for which we assess exact-match accuracy, and 949 open-ended questions, evaluated using recall, defined as the ratio of ground truth tokens present in the prediction.

\label{appen:perturbation_methods}
\begin{table}[h]
    \centering
    \includegraphics[width=0.28\textwidth]{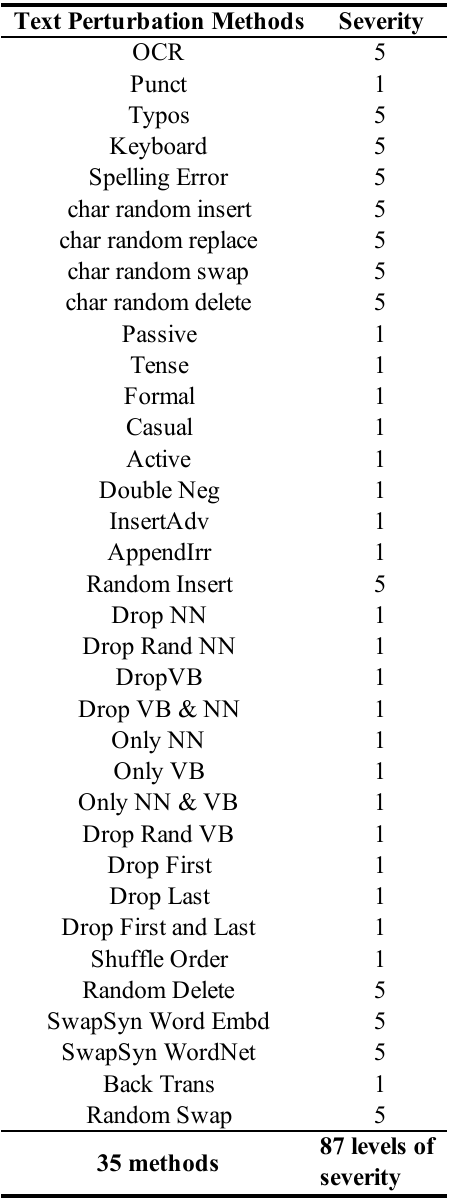}
    \caption{\label{text_perturbation_methods} Summary of text perturbation methods introduced by \citet{perturbation}.}
\end{table}

\begin{table}[h]
    \centering
    \includegraphics[width=0.28\textwidth]{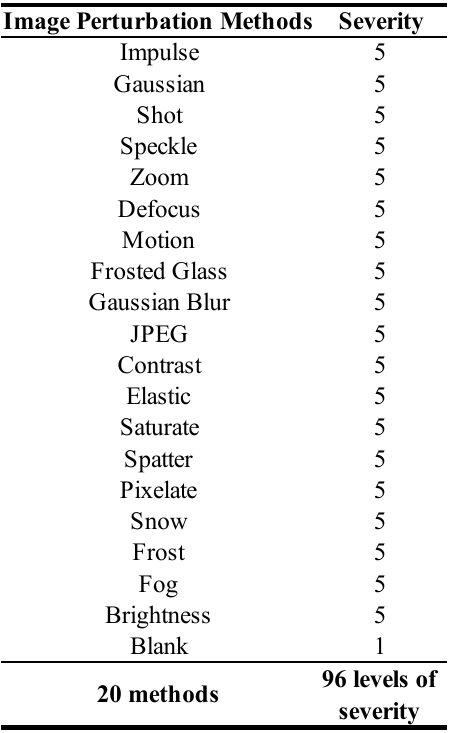}
    \caption{\label{image_perturbation_methods} Summary of image perturbation methods introduced by \citet{perturbation}.}
\end{table}

\section{Hyper-Parameter Settings}
\label{appen:hyper-param}
For all the token reduction methods, we choose the signature or best-performing hyper-parameter settings as reported in the corresponding papers.Specifically, for VisionZip \citep{visionzip}, the number of retained tokens was set to 192. For PyramidDrop \citep{pyramidDrop}, we use pruning layers at indices [8, 16, 24] and corresponding pruning ratios of [0.5, 0.25, 0.125]. For SparseVLM\citep{SparseVLM}, the number of retained tokens is set to 192. For FastV\citep{FastV}, we utilize settings of K=3 and R=0.5. Finally, HiRed\citep{Hired} was configured with a token budget of 20\%. These settings were consistently applied across relevant experiments.

\section{Input Perturbation Details}

To evaluate robustness, we utilize a comprehensive suite of input perturbation techniques proposed by \citet{perturbation}. The specifics of these perturbations are detailed for text in Table \ref{text_perturbation_methods} and for images in Table \ref{image_perturbation_methods}. Accounting for various severity levels, these amount to 87 distinct configurations for text inputs and 96 for image inputs. We randomly apply these varied perturbations to the text and image inputs of the VQAv2 \citep{balanced_vqa_v2} and GQA\citep{gqa} datasets. Importantly, to ensure a fair and consistent comparison across experiments, the exact same perturbed inputs are used for all tested acceleration methods.

\begin{figure*}[h]
\centering
\begin{subfigure}{0.49\textwidth}
  \includegraphics[width=\linewidth]{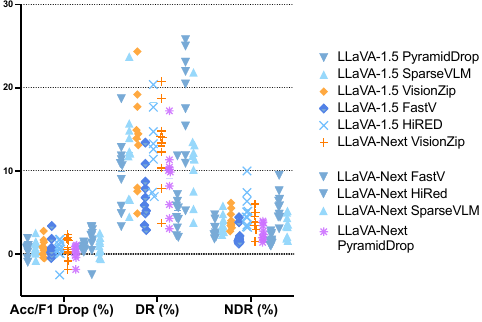}
  \caption{Token reduction distributions.}
  \label{pic:token_reduction_distri}
\end{subfigure}\hfill
\begin{subfigure}{0.49\textwidth}
  \includegraphics[width=\linewidth]{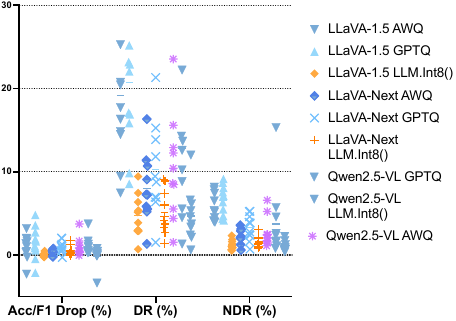}
  \caption{Quantisation distributions.}
  \label{fig:quant_distribution}
\end{subfigure}
\caption{Distribution of Acc/F1 drop vs.\ DR/NDR for (a) token reduction and (b) quantisation.}
\label{fig:distributions}
\end{figure*}


\begin{figure}[h]
\centering
\includegraphics[width=0.48\textwidth]{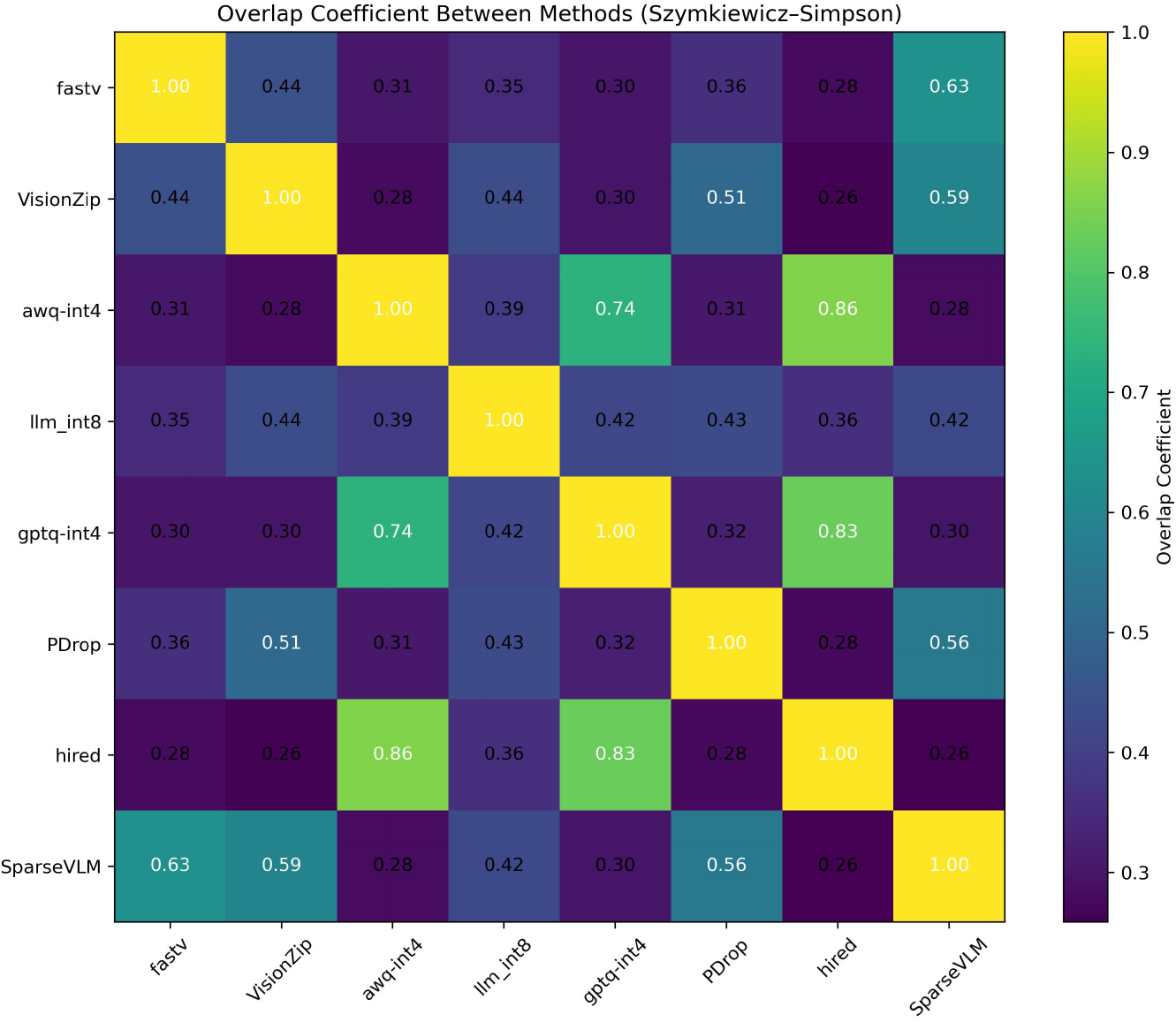}
\caption{\label{pic:overlapping} Overlap ratios of negatively diverged instances among acceleration methods for LLaVA-1.5 \citep{llava1.5}.}
\end{figure}

\begin{table*}[h]
\centering
\includegraphics[width=0.9\textwidth]{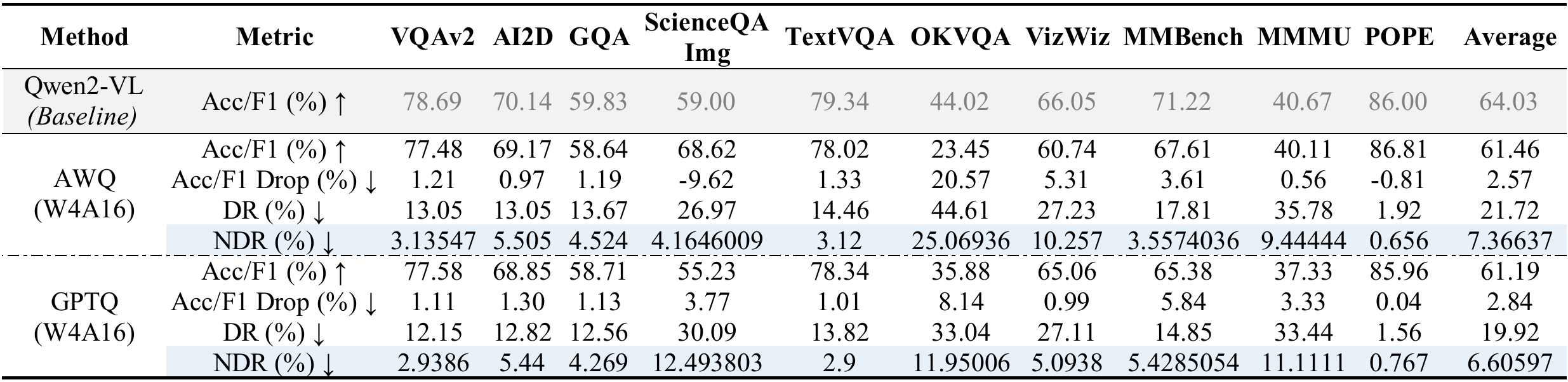}
\caption{\label{tab:qwen2vl} Instance-Level Instability of quantisation methods \citep{awq,gptq} in Qwen2-VL model \citep{Qwen2-vl}.}
\end{table*}

\section{Acceleration Methods Divergence Direction}
We further investigate the "divergence direction" of acceleration methods by examining the extent to which they are affected by the same instances. A high degree of overlap in these instances suggests that different methods diverge in a predictable, controllable manner. This shared divergence would simplify the development of universal solutions to mitigate instability. Conversely, minimal overlap—indicating highly separated divergences—would imply more unpredictable behavior, posing greater uncertainty for the practical deployment of these methods. To explore this, we analyzed results from LLaVA-1.5 \citep{llava1.5}, measuring the overlap of affected instances across various acceleration techniques. The findings are presented in Figure \ref{pic:overlapping}, which demonstrates that most pairings exhibit more "highly separated" divergences.

\section{Data Visualisation}
\label{appen:data_visual}
To better view the distribution of Acc/F1 Loss together with DR and NDR values, we plot a scatter diagram for Token Reduction Methods and Quantisation Methods, respectively. As shown in figure \ref{fig:distributions}, it reveals a consistent trend across various models and methods. In both diagrams, the "Acc/F1 Drop (\%)" remains notably low, generally appearing under 5\% and often close to or below 2\%. In stark contrast, the "DR (\%)" and "NDR (\%)" values are substantially higher, frequently ranging between 10\% and 30\%. This significant disparity underscores that while the accuracy or F1 score experiences minimal degradation, the other metrics, DR and NDR, show much more pronounced changes.

\section{Qwen2-VL Results}
\label{appen:qwen2-vl}
We conduct experimetns on Qwen2-VL \citep{Qwen2-vl} 3B model with AWQ\citep{awq} and GPTQ\citep{gptq} quantisation methods, detailed in table \ref{tab:qwen2vl}. It reveals varied performance impacts across different benchmarks. On average, AWQ quantization leads to a 2.57\% drop in Acc/F1 score, an outcome notably influenced by an unexpected 9.62\% performance increase on the ScienceQA Img benchmark, alongside a significant 20.57\% performance decrease on OKVQA. GPTQ quantization results in a slightly higher average Acc/F1 drop of 2.84\%, with its most pronounced performance reductions observed on OKVQA (8.14\% drop) and MMBench (5.84\% drop). While the average changes in Acc/F1 scores are relatively contained, both quantization techniques generally cause substantial increases in DR (\%) and NDR (\%) values across the evaluated benchmarks.

\begin{table}[h]
\centering
\includegraphics[width=0.48\textwidth]{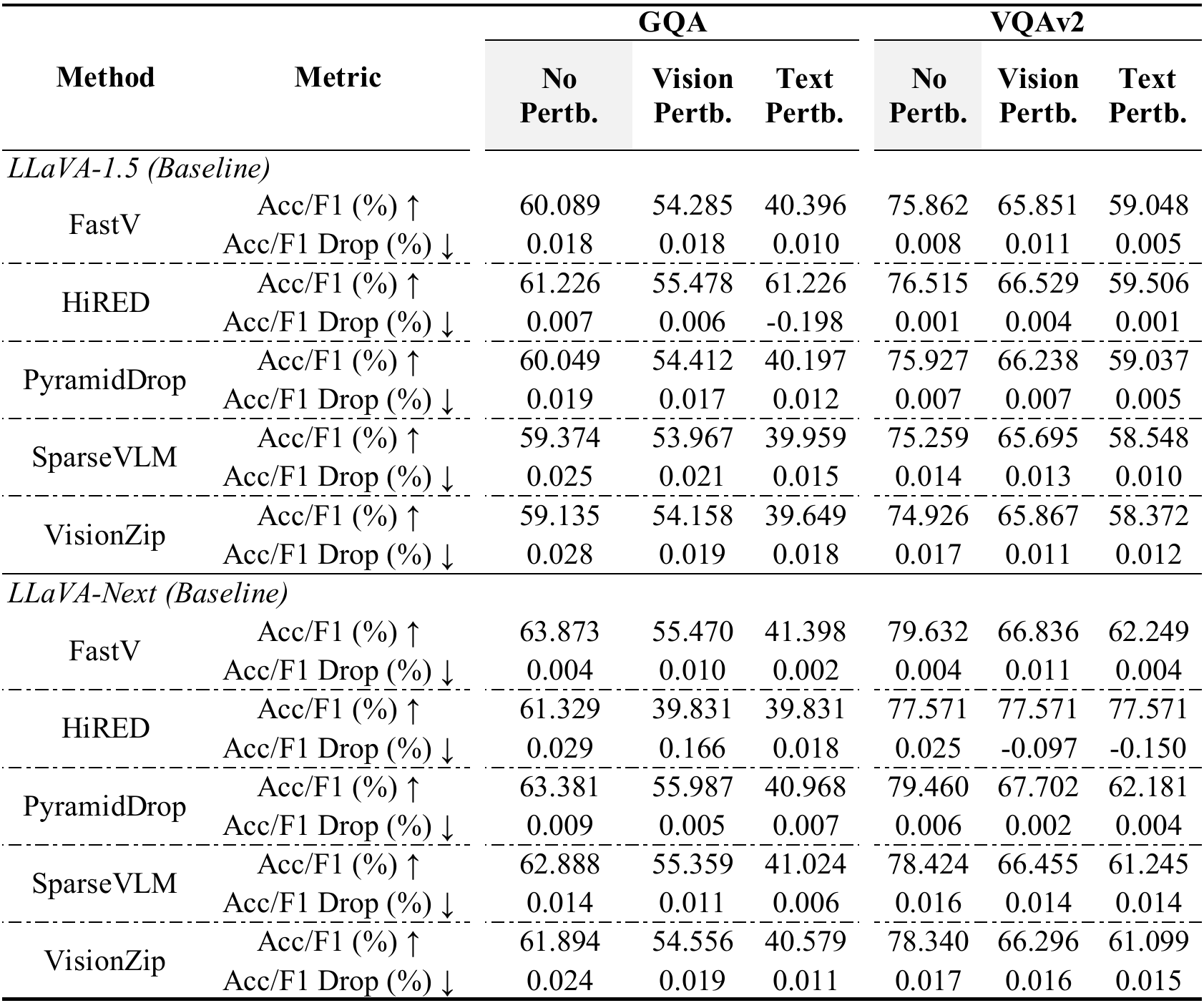}
\caption{\label{tab:acc/f1_tokenred} Performance and performance drop of token reduction methods under input perturbation.}
\end{table}
\begin{table}[h]
\centering
\includegraphics[width=0.48\textwidth]{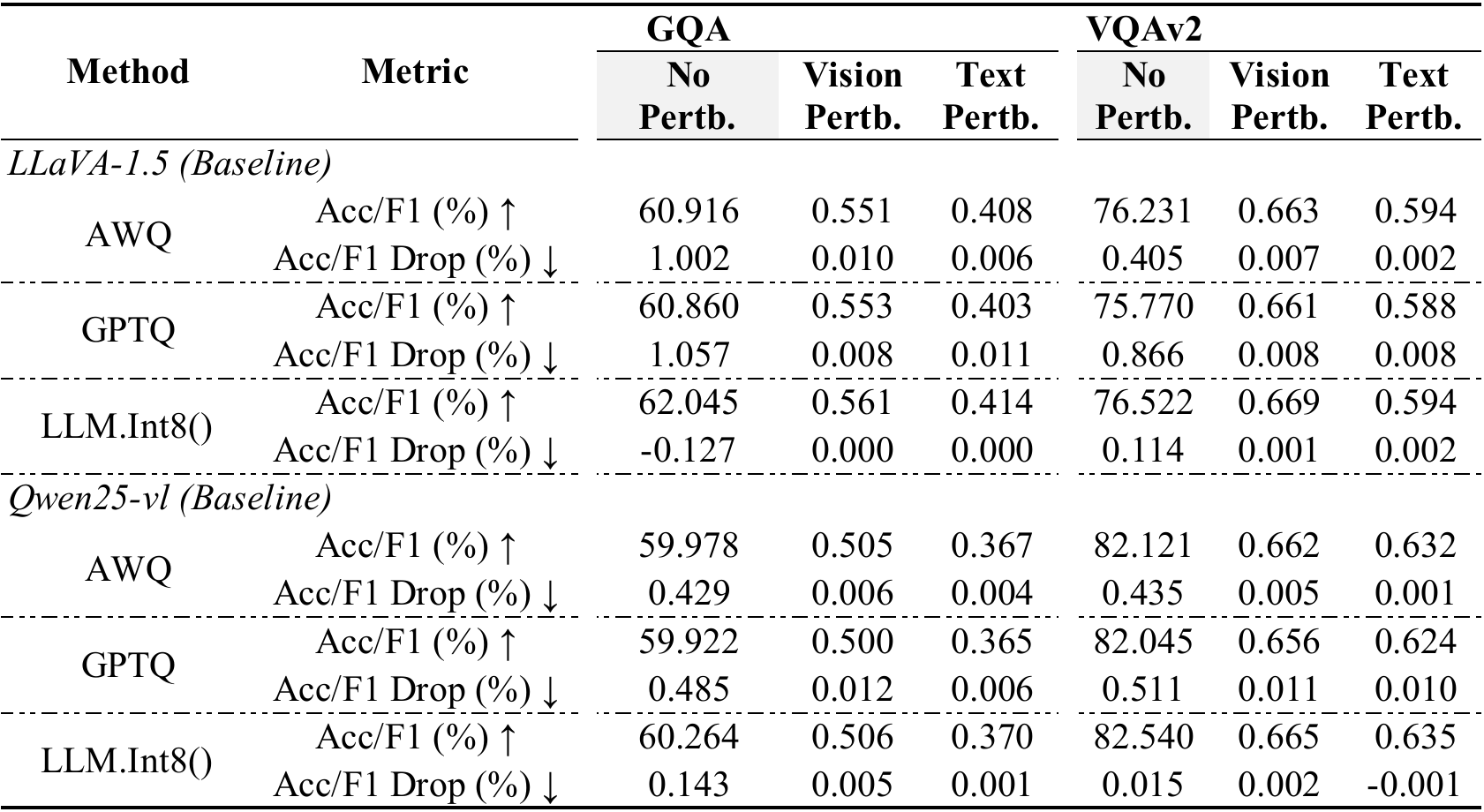}
\caption{\label{tab:acc/f1_quant} Performance and performance drop of quantisation methods under input perturbation.}
\end{table}

\section{Input Perturbation Impacts on Acc/F1 and Acc/F1 Drop}
Table \ref{tab:acc/f1_tokenred} and table \ref{tab:acc/f1_quant} detail the performance of various acceleration techniques—quantization (AWQ, GPTQ, LLM.Int8()) and token reduction (FastV, HIRED, PyramidDrop, SparseVLM, VisionZip)—on models like LLaVA-1.5, LLaVA-Next, and Qwen2.5-vl, across GQA and VQAv2 datasets under no, vision, and text perturbations. A consistent trend across both sets of methods is the remarkably low impact on Acc/F1 scores; the Acc/F1 Drop (\%) is generally minimal, often well below 1\% and frequently in the hundredths of a percent, irrespective of the specific acceleration technique or perturbation type applied. 

\section{Qualifying Examples}
\label{appen:qualifying_examples}
In this section, we present qualifying examples: specific test instances showing how applying acceleration methods to a Vision Language Model (VLM) can cause prediction divergence.
\begin{figure}[ht]
\centering
\includegraphics[width=0.47\textwidth]{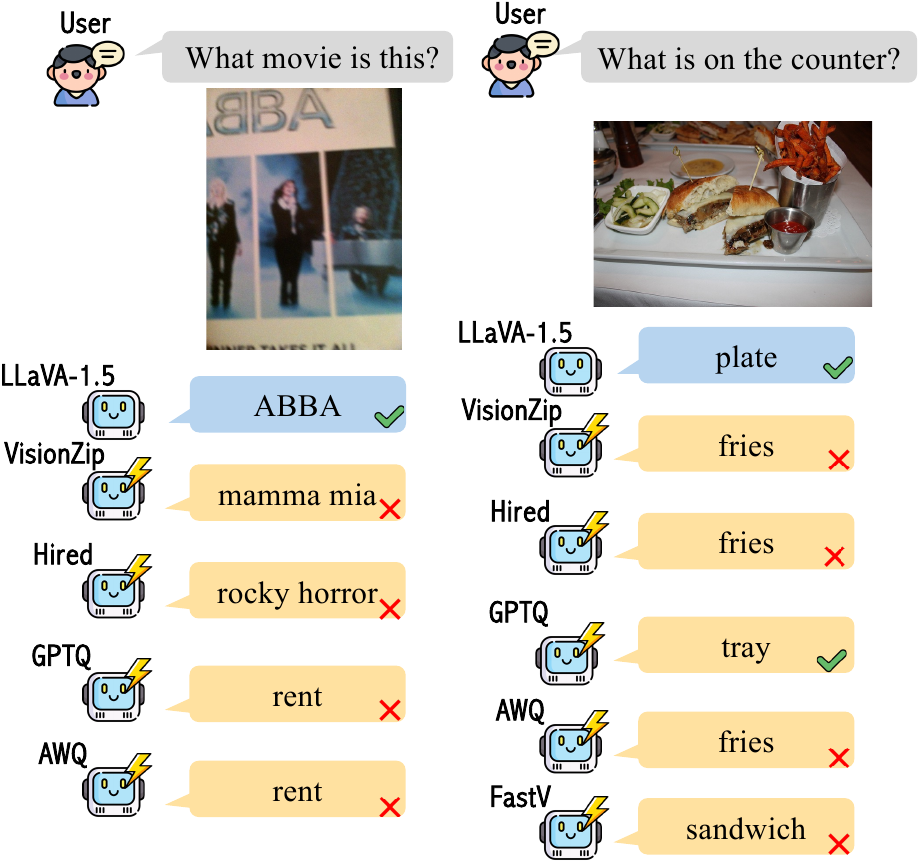}
\caption{Acceleration Instances Divergence qualifying examples for LLaVA-1.5 \citep{llava1.5}.}
\end{figure}
\begin{figure}[ht]
\centering
\includegraphics[width=0.47\textwidth]{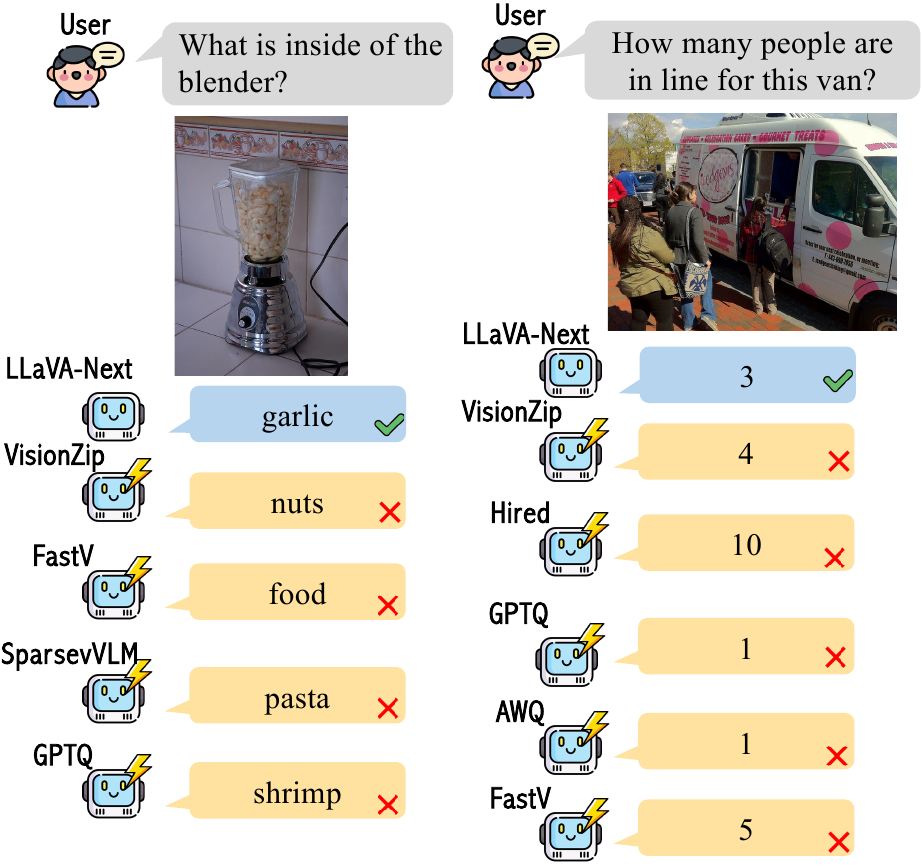}
\caption{Acceleration Instances Divergence qualifying examples for LLaVA-Next \citep{liu2024llavanext}.}
\end{figure}
\begin{figure}[ht]
\centering
\includegraphics[width=0.47\textwidth]{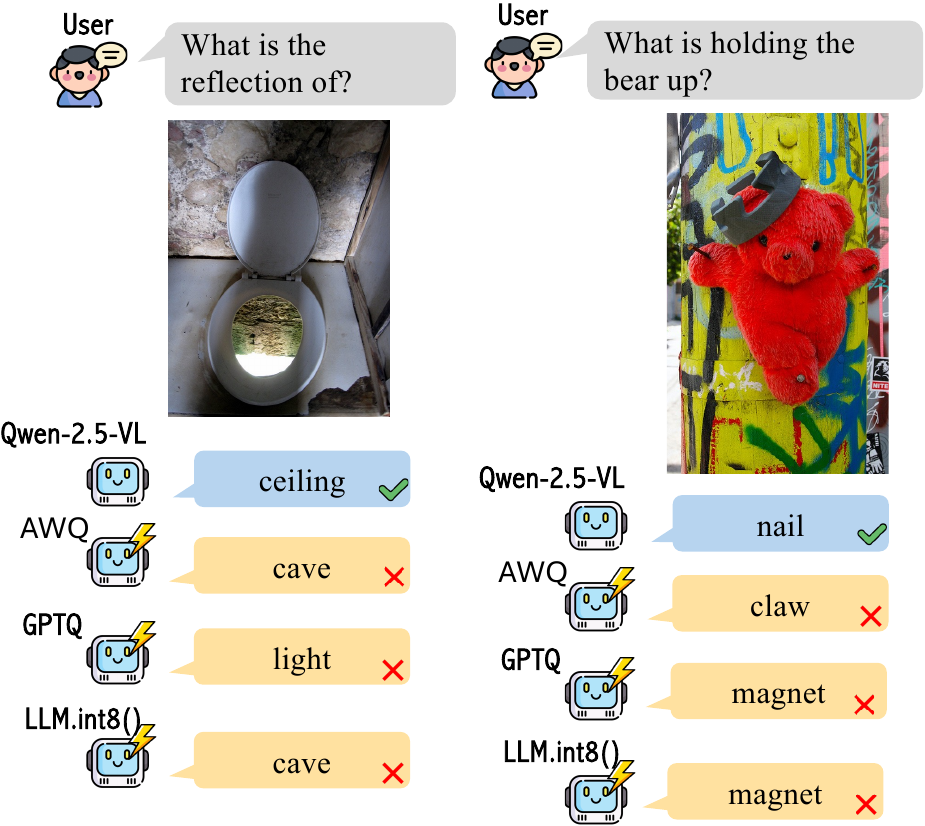}
\caption{Acceleration Instances Divergence qualifying examples for Qwen2.5-VL \citep{qwen2.5-vl}.}
\end{figure}
\begin{figure}[ht]
\centering
\includegraphics[width=0.47\textwidth]{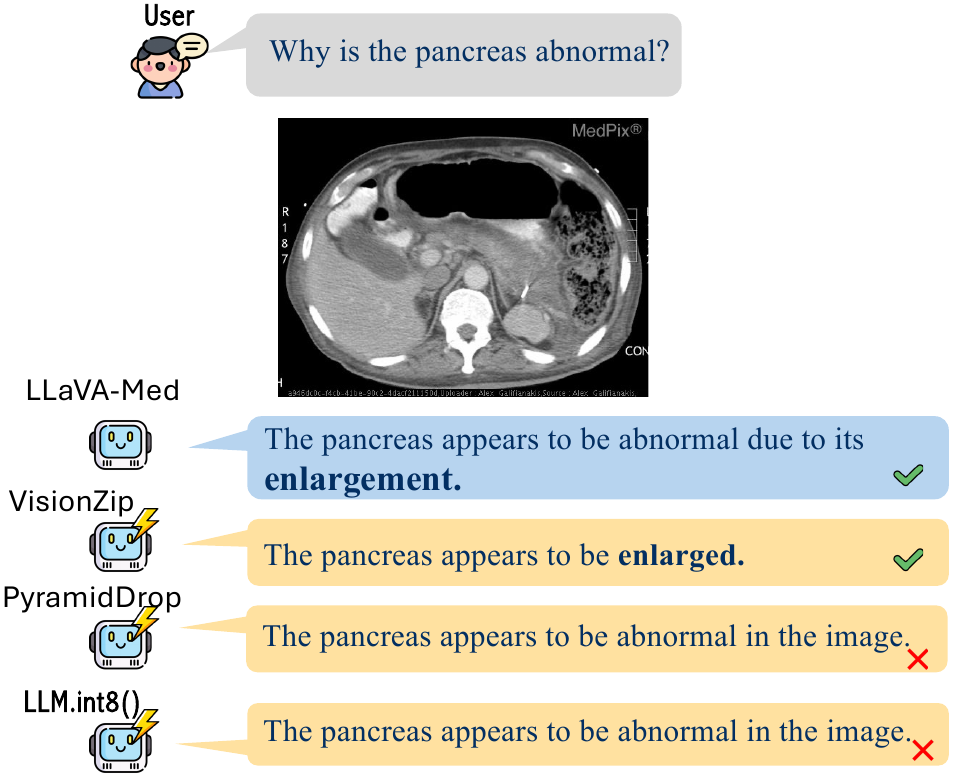}
\caption{Acceleration Instances Divergence qualifying examples for LLaVA-Med \citep{llava-med}.}
\end{figure}
\begin{figure}[ht]
\centering
\includegraphics[width=0.47\textwidth]{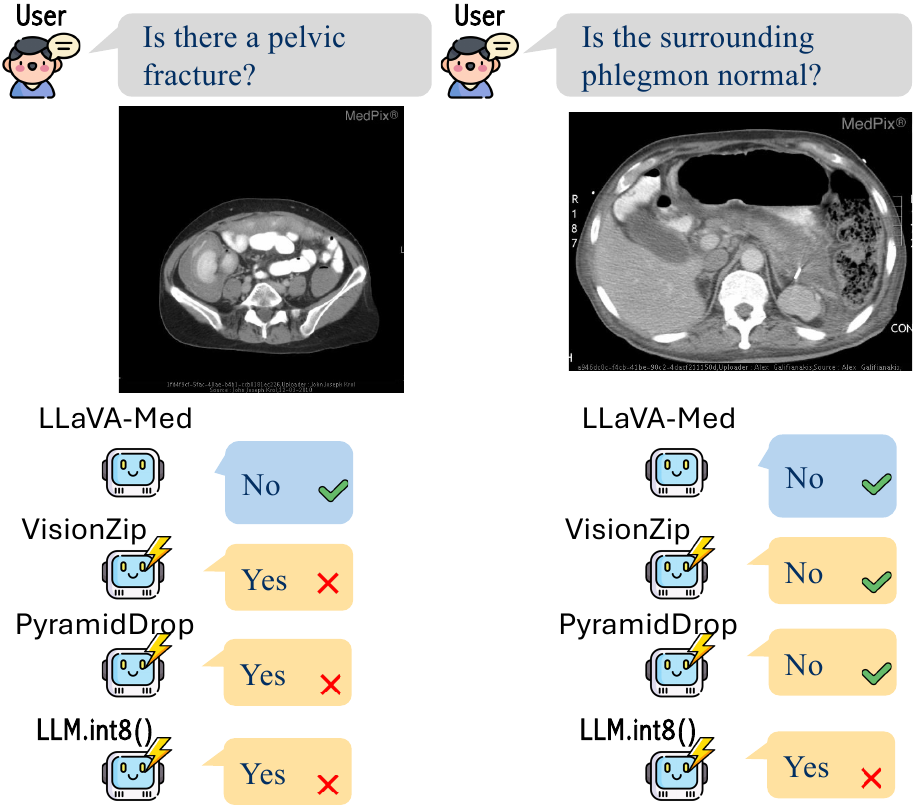}
\caption{Acceleration Instances Divergence qualifying examples for LLaVA-Med \citep{llava-med}.}
\end{figure}
\end{document}